\documentclass[runningheads]{llncs}
\usepackage[T1]{fontenc}
\usepackage{graphicx}
\usepackage{multirow}
\usepackage{booktabs}
\usepackage{bbding}   
\usepackage{colortbl}  
\usepackage{makecell}
\usepackage{amssymb}
\usepackage{utfsym}
\usepackage{amsmath}
\usepackage{xcolor}
\usepackage{cancel}
\usepackage{tabularx}
\usepackage{graphicx}
\usepackage{amsfonts}
\usepackage{orcidlink}
\makeatletter
\def\@oddfoot{\hfill} 
\let\@evenfoot\@oddfoot
\makeatother

\begin{document}
\title{Ignite Forecasting with \textit{SPARK}: An Efficient Generative Framework for Refining LLMs in Temporal Knowledge Graph Forecasting}
\titlerunning{SPARK: A Generative Framework for Refining LLMs in TKG Forecasting}
%
%
\author{Gongzhu Yin\inst{1}\orcidlink{0000-0002-8251-742X} \and
Hongli Zhang*\inst{1} \and
Yi Luo\inst{1} \and
Yuchen Yang\inst{1} \and
Kun Lu\inst{1} \and
Chao Meng\inst{1}}
\authorrunning{G. Yin et al.}
%
\institute{School of Cyberspace Science, Harbin Institute of Technology, Harbin, China \\
\email{yingz@hit.edu.cn, zhanghongli@hit.edu.cn, yi\_luo@stu.hit.edu.cn, yangyc@hit.edu.cn, lukun@hit.edu.cn, mengchao@hit.edu.cn}}

\maketitle              
\begin{abstract}
Temporal Knowledge Graph (TKG) forecasting is crucial for predicting future events using historical data. With the surge of Large Language Models (LLMs), recent studies have begun exploring their integration into TKG forecasting and achieved some success. However, they still face limitations such as limited input length, inefficient output generation, and resource-intensive refinement, which undermine their performance and practical applicability. To address these limitations, we introduce \textit{SPARK}, a \textbf{S}equence-level \textbf{P}roxy-\textbf{A}dapting framework for \textbf{R}efining LLMs in T\textbf{K}G forecasting. Inspired by inference-time algorithms adopted in controlling generation, SPARK offers a cost-effective, plug-and-play solution through two key innovations: (1) Beam Sequence-Level Generation, which reframes TKG forecasting as a top-K sequence-level generation task, using beam search for efficiently generating next-entity distribution in a single forward pass. (2) TKG Adapter for Refinement, which employs traditional TKG models as trainable proxy adapters to leverage global graph information and refine LLM outputs, overcoming both the input length and the resource-intensive fine-tuning problems. Experiments across diverse datasets validate SPARK's forecasting performance, robust generalization capabilities, and high efficiency. We release source codes at \textcolor{blue}{{\url{https://github.com/yin-gz/SPARK}}}.

\end{abstract}
\section{Introduction}
Reasoning about future events based on historical information is an active research area with wide-ranging applications. To model large amounts of real-world event data, temporal knowledge graphs (TKGs) have been introduced. TKGs store temporal events as a series of timestamped quadruples ${(e_s,r,e_o,t)}$, where $e_s$ and $e_o$ interact via relation $r$ at time $t$. TKG forecasting aims to predict missing entities in ${(e_s,r,?,t)}$ given historical events before $t$.

Traditional TKG forecasting relies on temporal rule mining \cite{TLogic} or supervised deep learning, such as embedding learning \cite{CEN} and graph neural networks (GNNs) \cite{REGCN}. However, they require dataset-specific training and lack generalization. With LLMs' rise, recent studies \cite{ICLTKG,gentkg,onsep,LLMRule} have begun exploring their integration into TKG forecasting, leveraging their knowledge and generative abilities. Most of them follow an in-context learning (ICL) framework \cite{ICLTKG}, extracting relevant historical events, formatting them as temporally ordered quadruples, and prompting LLMs to predict next possible entities.

\begin{figure*}\centering
	\includegraphics[width=0.99\linewidth]{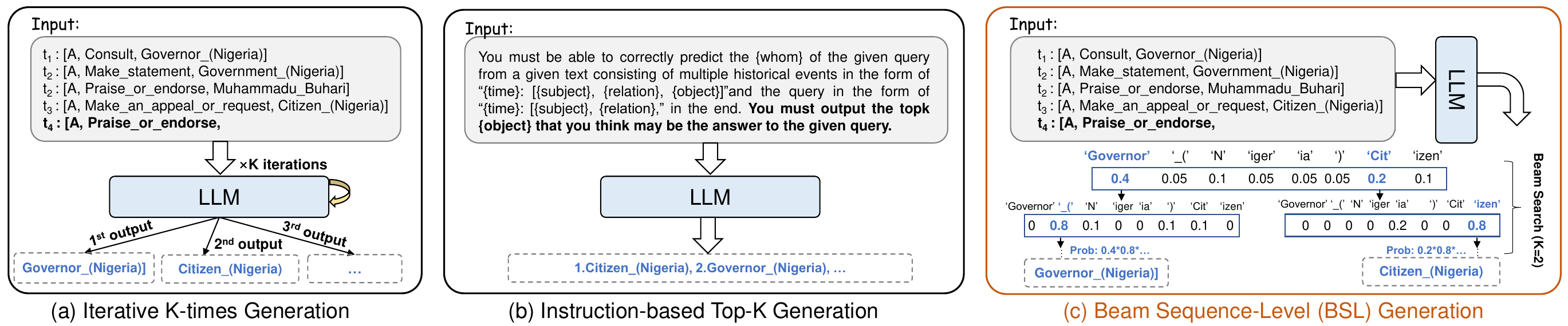}
	\caption{Examples comparing the proposed BSL Generation with iterative K-times generation \cite{gentkg} and instruction-based top-K generation \cite{CoHTKG}. Figure (c) takes the top-2 as an example, where two tokens with the highest probabilities are selected at each step.}
	\label{fig_example}
\end{figure*}

Despite advancements, LLM-based methods still face key limitations: \textbf{(1) Limited Input Length.} The finite context window of LLMs makes it challenging to include all historical information and graph structures from a TKG, risking crucial entity omissions and inaccurate predictions. \textbf{(2) Inefficient Output Generation.} Generating top-K entities conflicts with LLMs’ next-token paradigm, often requiring iterative K-times runs \cite{gentkg} or handcrafted prompts \cite{CoHTKG} to instruct LLMs to list top-K entities, increasing inefficiency (as shown in Figure \ref{fig_example}). \textbf{(3) Resource-Intensive Task Alignment.} To further align LLMs with TKG forecasting tasks, several studies \cite{gentkg} have employed instruction-tunning (IT), which demands high computational cost and time.

To align LLMs with task-specific needs at lower costs, inference-time algorithms for controlling generation have gained significant attention \cite{IPA}. These algorithms tailor the model’s next token distribution by refining it with external adapters, leaving the underlying LLM unchanged. Inspired by this, we introduce SPARK, a \textbf{S}equence-level \textbf{P}roxy \textbf{A}dapting framework for \textbf{R}efinng LLMs in T\textbf{K}G forecasting. SPARK adapts inference-time algorithms to TKG forecasting through two key innovations:
\textbf{(1) Beam Sequence-Level (BSL) Generation:}
Reformulate entity prediction as sequence-level (N-step tokens) generation, leveraging beam search for instruction-free top-K entity generation in a single pass, addressing output inefficiencies.
\textbf{(2) TKG Adapter for Refinement:}
Employ traditional TKG models as proxy adapters to refine LLM’s sequence-level output, reducing fine-tuning costs while integrating structural reasoning, effectively addressing both the input length limitations and the resource-intensive fine-tuning. Through this process, the adapters learn to identify what kinds of relations the LLM handles well and which depend more on structural information. To validate SPARK, we conduct extensive experiments on single-step TKG forecasting to assess its accuracy and generalization. Results show that SPARK consistently enhances LLM performance, achieving competitive performance and generalization with IT-tuned models while boosting efficiency.

\section{Preliminaries}
\subsection{Problem Definition of TKG Forecasting}
Let $\mathcal{G}=\{\mathcal{E},\mathcal{R},\mathcal{T},\mathcal{Q}\}$ denote a TKG, where $\mathcal{E}$, $\mathcal{R}$, $\mathcal{T}$ are the entity set, the relation set, the timestamp set, respectively. Each fact is denoted as a quadrupled $(e_s,r,e_o,t) \in \mathcal{Q}$. Moreover, since timestamp is sequential, $\mathcal{G}$ can be split into a time-sequenced series of multi-relational directed graphs, $\mathcal{G} = \{\mathcal{G}_1,\mathcal{G}_2,...,\mathcal{G}_t,...\}$, where each $\mathcal{G}_t$ contains facts occuring at time $t$. Given a query quadruple $q=(e_s,r,?,t)$, the TKG forecasting task aims to predict the missing entity based on historical KG sequence $\mathcal{G}_{<t} = \{\mathcal{G}_1,\mathcal{G}_2...,\mathcal{G}_{t-1}\}$.

In LLM-based methods, the limited input length makes them not feasible to consider all candidate entities in $\mathcal{E}$. Instead, they generate only the top-K possible answers using iterative K-times generation \cite{gentkg} with varying temperature settings or constraint instructions \cite{CoHTKG,onsep} like "You must output the top-K answers".

\subsection{Refining LLMs at Inference Time}
To refine LLMs for specific tasks or requirements, \cite{IPA} introduced an inference-time adaptation framework called IPA. When generating the next token at step $t$ based on the previously generated $t-1$ tokens, IPA combines the output distributions of a lightweight adapting network and the frozen LLM as follows:
\begin{equation}
P_{\theta \leftarrow \phi}\left(\boldsymbol{y}_t|\boldsymbol{y}_{<t}\right) \propto P_\theta\left(\boldsymbol{y}_t|\boldsymbol{y}_{<t}\right) P_\phi\left(\boldsymbol{y}_t|\boldsymbol{y}_{<t}\right)
\end{equation}
where $\boldsymbol{y_t}$ denotes the token generated at $t$ step, $P_{\theta}$ and $P_{\phi}$ denote the output distributions of the LLM and the adapting model, respectively. $P_{\theta \leftarrow \phi}$ represents the refined output distribution. During fine-tuning, only the parameters of the adapter ($\phi$) are updated, which can be much smaller than those of the base LLM ($ \theta $). Inference-time algorithms often use smaller LMs as proxy adapters. However, for TKG tasks, they still face the limited input problem. 

\section{The Proposed SPARK}
\begin{figure*}\centering
	\includegraphics[width=0.9\linewidth]{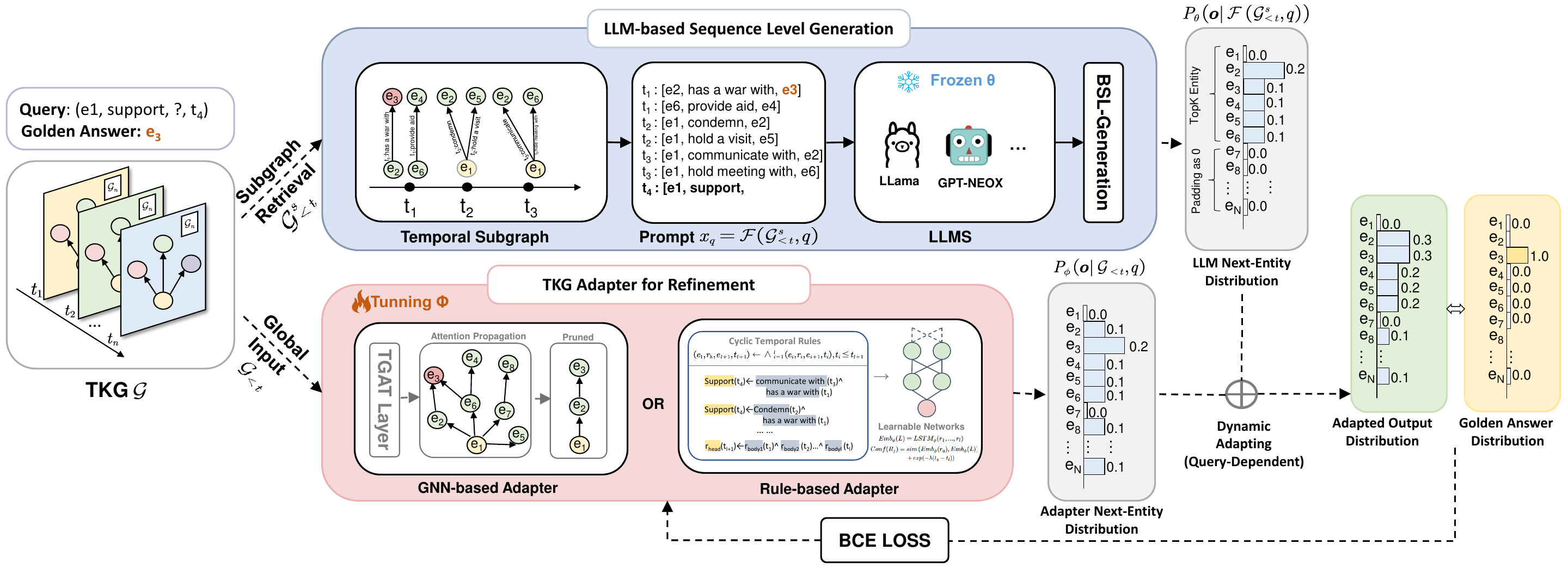}
	\caption{The overall structure of SPARK.}
	\label{fig_main}
\end{figure*}

An overview of SPARK is shown in Figure \ref{fig_main}. In the first stage, the LLM generates the next entity distribution based on retrieved historical sequences \textbf{(Sec \ref{sec:gen})}. Simultaneously, the adapting models operate on the global graph, learning temporal patterns and producing their own next entity distribution \textbf{(Sec \ref{sec:adapt})}. We can see that candidate entities like "e7" and "e8," omitted by the LLM due to input length limitations, are considered by the TKG Adapter. In the next stage, these two distributions are dynamically combined, resulting in an adapted output distribution as the final prediction. Note that, only the adapting models are updated during training, making the refining cost-effective \textbf{(Sec \ref{sec:output})}.

\subsection{LLM-based Sequence Level Generation}
\label{sec:gen}
To leverage LLMs to generate the next entity distribution, we reframe TKG forecasting as sequence-level generation and employ a beam search decoding strategy to retrieve the next N tokens as a sequence with top-K probabilities.

\subsubsection{Prompt Construction}
When constructing the input prompt for the LLM, we follow the widely used ICL framework. For a given query $q=(e_q,r_q,?,t_q)$ and historical TKG $\mathcal{G}_{<t}$, we initially retrieve a subset of relevant historical contexts $\mathcal{G}_{<t}^s$. SPARK framework is compatible with arbitrary retrieval methods, enabling integration with key-based \cite{ICLTKG}, TLogic-based \cite{gentkg} or causal-based \cite{onsep}.

The prompt is then constructed by transforming historical contexts into a sequence of temporally ordered TKG quadruples, with the query appended at the end, denoted as $x_q = \mathcal{F}(\mathcal{G}_{<t}^s, q)$. The transforming function $\mathcal{F}$ operate as follows: 
\textbf{(1)} For each historical quadruple $(e_{s_i},r_i,e_{o_i},t_i) \in \mathcal{G}_{<t}^s$, it is represented as a textual sequence "$t_i:[(e_{s_i},r_i,e_{o_i})]$". 
\textbf{(2)} The query is formatted similarly as "$t_q:[e_{q},r_q,$", appended at the end.

This structure enables LLMs to generate the next entity naturally. Entities and relations can be represented in indexed (e.g., 1024) or lexical (e.g., Citizen\_(Nigeria)) format. Previous studies \cite{ICLTKG} have shown that both formats yield similar accuracy, but the indexed format significantly reduces prompt consumption. In practice, we represent relations in their lexical form, such as 'Accuse', to leverage the LLM’s internal knowledge. Conversely, entities are denoted by indexed identifiers to minimize prompt length and prevent potential data leakage. Note that, each integer (i.e. "0-10") is treated as a single token. For example, the entity "1024" is split into multiple tokens: "1", "0", "2", and "4".

\subsubsection{Beam Sequence-Level (BSL) Generation}
Since entities may consist of multiple tokens, we cast TKG forecasting as a sequence-level generation task and use a beam search strategy to efficiently decode the next entity distribution in a single forward pass (as shown in Figure \ref{fig_example} (c)). Beam search generates a set of candidate sequences and selects the top-K with the highest probabilities:
\begin{equation}
P_\theta\left(\left\{y_t^{(i)}\right\}_{i=1}^N\right)=\prod_{i=1}^N P_\theta\left(y_t^{(i)} \mid x_q, y_t^{1: i-1}\right)
\end{equation}
where $y_t^{(n)}$ is the $n$-th token in the generated sequence at timestamp $t$. The beam search algorithm proceeds until the sequence is complete, either by reaching the maximum length or generating a stop token "]". Compared to iterative K-times generation \cite{gentkg} or instruction-based methods \cite{CoHTKG,onsep}, beam search is more efficient, reduces computational overhead, and directly outputs the probability for each sequence—crucial for integrating with the adapter framework. At this stage, it is compatible with most open-source models but may not work with certain API-based models like ChatGPT.

\subsubsection{Output Next-Entity Distribution}
The top-K sequences from the LLM are mapped to entities using the entity mapping function $\mathcal{M}_{ent}$. It operates as follows: If the output is a valid entity index, it converts the index string to an integer. For invalid outputs, such as out-of-vocabulary tokens or unexpected answers, the outputs are simply discarded.
Next, we apply a softmax to the top-K predicted entities, setting the probabilities of all other entities to zero to form a complete entity distribution. Formally, the next-entity output distribution $P_\theta(\boldsymbol{o} \mid \mathcal{F}(\mathcal{G}_{<t}^s, q))$ is defined as:
\begin{equation}
P_\theta(\boldsymbol{o}) = \text{softmax}\left( \{ P_\theta(e_{a_1}), \dots, P_\theta(e_{a_K}) \} \right) \cup \{ 0 \text{ for all other entities} \} 
\end{equation}
where $e_{a_*} \in \mathcal{A}_{LLM}^q$ and $P_\theta(e_{a_*})$ are the predicted probability assigned to the entity $e_{a_*}$. If no entities can be extracted, we set the output as a zero distribution.

\subsection{TKG Adapter for Refinement}
\label{sec:adapt}
To refine the LLM's next-entity output, we integrate traditional TKG forecasting models as adapters within a side-tuning framework. These adapters, with adjustable parameters $\phi$, take the entire historical graph as input and compute the next-entity distribution, represented as $P_{\phi}(\boldsymbol{o}|\mathcal{G}_{<t}, q)$. This distribution is then used to refine the LLM's output. While LLMs typically focus on local information and textual correlations, these adapters integrate multi-hop reasoning and logical rules, providing more context-aware information and complementing the LLM’s structural reasoning capabilities. We explored the following two types of traditional TKG forecasting adapters.

\subsubsection{GNN-based Adapter}
Given that a TKG is a sequence of graphs, GNNs are widely used in traditional TKG forecasting methods when combing with sequence models \cite{REGCN,cygnet}. Unlike LLMs, GNN-based methods excel at capturing temporal patterns, providing complementary strengths to LLMs' text-based reasoning, making them ideal adapters for enhancing TKG forecasting. Among the available GNN-based methods, we implement xERTE \cite{XERTE}, an explainable GNN that expands and prunes subgraphs hop-wise, offering both strong performance and interpretability by revealing influential nodes in LLM integration.

\subsubsection{Rule-based Adapter}
In addition to GNN-based models, another prominent line of work employs symbolic frameworks to learn explicit temporal logical rules. One pioneering approach, TLogic \cite{TLogic}, extracts temporal walks from graphs and generalize cyclic temporal rules of form: $(e_1,r_h,e_{l+1},t_{l+1}) \leftarrow \wedge_{i=1}^l\left(e_i, r_i, e_{i+1}, t_i\right)$, $t_1 ... \leq t_{l} \leq t_{l+1}$. Recently, neuro-symbolic methods have emerged, combining neural embeddings with explicit rule learning and inference, blending the interpretability of symbolic models with the parameter-sharing capabilities of neural networks \cite{NeuSTIP}. Inspired by neuro-symbolic approaches, we adapt TLogic to function as an adapter for LLMs. Specifically, we make the weights of each mined rule learnable. For the query $(e_q, r_q, ? ,t_q)$ with a rule $R_j$ having a rule head $r_q$ and a rule body $L = (r_1,...,r_l)$, the confidence of rule $R_j$ is computed as:
\begin{equation}
Emb_{\phi}(L) = LSTM_{\phi} (r_1, ..., r_l)
\end{equation}
\begin{equation}
\begin{aligned}
Conf(R_j) = sim\left(Emb_{\phi}(r_q), Emb_{\phi}(L)\right) + exp(-\lambda(t_q-t_l))
\end{aligned}
\end{equation}
where $\lambda>0$ is a hyperparamter. By traversing all corresponding rules, we obtain the candidate entity set $\mathcal{A}_{Ada}^q$ and the sum of their confidences. As described in Sec \ref{sec:gen}, we then generate the next-entity distribution of the adapter.

\subsection{Dynamic Adapting and Training}
\label{sec:output}
Following inference-time algorithms, we refine the next-entity distribution of LLM using the TKG adapter, denoted as:
\begin{equation}
P_{\theta \leftarrow \phi} (\boldsymbol{o}|\mathcal{G}_{<t}, q) \propto P_{\theta}(\boldsymbol{o}|\mathcal{F}(\mathcal{G}_{<t}^s, q))
P_{\phi}(\boldsymbol{o}|\mathcal{G}_{<t}, q)
\end{equation}

Unlike static model fusion, our approach dynamically weights the adapter distribution based on the query. For GNN-based adapter, the weight of the adapter distribution is calculated using MLPs with a Sigmoid activation function, where the learned query relation embeddings serve as the input. For rule-based adapter, we directly average the two distributions as the adapter internally calculates query-dependent rule weights. This adaptive weighting allows the model to balance LLM reasoning with structural information. Finally, we train the adapter parameter $\phi$ using a binary cross-entropy (BCE) loss:
\begin{equation}
	\mathcal{L} = \sum_{j=1}^{|\mathcal{E}|}{y_j\log}P_{\theta \leftarrow \phi} (o_j|\mathcal{G}_{<t}, q)-\left( 1-y_j \right) \log \left[ 1-P_{\theta \leftarrow \phi} (o_j|\mathcal{G}_{<t}, q) \right]
\end{equation}
where $o_j$ is a candidate, with $y_j=1$ for the true target and $y_j=0$ otherwise.

\section{Experiments}
In the experiments, we delve into these research questions:
\textbf{(RQ1)} How do LLMs with SPARK compare to original LLMs, traditional TKG forecasting methods and LLMs with instruction tuning (IT) in terms of accuracy?
\textbf{(RQ2)} How does SPARK perform under inductive generalization scenes?
\textbf{(RQ3)} How efficient is SPARK compared to IT methods?
\textbf{(RQ4)} Ablation study for the two key innovations in SPARK.

\subsection{Experimental Settings
}
\subsubsection{Datasets}
To facilitate comparison with previous baselines (especially LLM-based), we evaluate SPARK on TKG forecasting using three widely adopted public datasets \cite{RENET}: ICEWS14, ICEWS18 , and GDELT. ICEWS14 and ICEWS18 are subsets extracted from the Integrated Crisis Early Warning System, which contain international daily events from 2014 and 2018, respectively. GDELT is a catalog of human societal-scale behavior extracted from news media. To train the adapter, we follow the same data split as in previous work \cite{gentkg} to ensure a fair comparison. The datasets are split into training, validation, and test sets, with the timestamps in each set occurring sequentially. The detailed statistics of the datasets are shown in Table \ref{tab:dataset}.

\begin{table}
    \centering
    \caption{The statistics of the datasets.}
    {\fontsize{7pt}{8pt}\selectfont
    \resizebox{0.99\textwidth}{!}{
    \begin{tabularx}{\textwidth}{c|X|X|X|X|X|c}
    \toprule
    \textbf{Datasets} & \# $\mathcal{E}$ & \# $\mathcal{R}$ & \# Train & \# Valid & \# Test & Interval \\
    \midrule
    ICEWS14  & 7,128 & 230 & 74,854 & 8,514 & 7,317 & 1 day \\
    ICEWS18  & 23,033 & 256 & 373,018 & 45,995 & 49,545 & 1 day \\
    GDELT    & 5,850 & 238 & 79,319 & 9,957 & 9,715 & 15 mins \\
    \bottomrule
    \end{tabularx}
    }}
    \label{tab:dataset}
\end{table}

\subsubsection{Baselines}
We evaluate SPARK against various baselines to measure its improvement over original LLMs and its performance compared to adapter-only and LLM-enhanced methods: \textbf{(1) Traditional TKG forecasting methods}, which include embedding-based models like xERTE \cite{XERTE} and RE-GCN \cite{REGCN}, as well as rule-based method like TLogic \cite{TLogic}. As our focus
is on measuring SPARK’s improvement over original LLMs, we omit other unrelated TKG methods. \textbf{(2) LLM-based methods.} When applied to TKG forecasting, LLM-based methods must be combined with retrieval strategies due to the limited input length. Two representative approaches are ICL (entity-key based) \cite{ICLTKG}  and TLR (rule-based) \cite{gentkg}, which are combined with LLama2-7B, GPT-NeoX-20B, and InterLM2 in previous studies with reported results. We also consider causal rule mining (InternLM2-7B-ONSEP) \cite{onsep}. Models like CoH \cite{CoHTKG} are excluded due to differing evaluation settings. We refer to SPARK combined with the GNN-based adapter as \textbf{SPARK(G)} and with the rule-based adapter as \textbf{SPARK(R)}. Since SPARK is plug-and-play, we integrate it into both LLM-ICL and LLM-TLR.

\subsubsection{Implementation Details}
We follow standard evaluation settings, including time-aware filtering, single-step prediction, and Hits@K metrics. When implementing, we use the efficient LLM inference framework vLLM to precompute the LLMs' output distributions and store the results. For each dataset, we set the training epochs to 5, the learning rate $lr$ to 1e-4, the batch size $\rho$ to 128. In all experiments, models are trained and tested three times with different random seeds, and the mean performance is reported.
\begin{table*}
	\centering
	\large
	\caption{Main performance comparison. $\boldsymbol{\circ}$ indicates training a traditional small model, \checkmark indicates training the LLM model, and \XSolidBrush indicates no extra training. $\triangle$ \textit{Improve} represents the percentage improvement over the original LLM. For all baselines, we use the reported results from previous studies \cite{gentkg} (iterative K-times generation) and \cite{onsep} (instruction-based top-K generation) conducted under the same settings.}
	\resizebox{\textwidth}{!}{
		\begin{tabular}{c|c|c|ccc|ccc|ccc}
			\toprule
			\multirow{2}[2]{*}{\textbf{Method Type}} & \multirow{2}[2]{*}{\textbf{Model}} & \multirow{2}[2]{*}{\textbf{Train}} & \multicolumn{3}{c|}{\textbf{ICEWS14}} & \multicolumn{3}{c|}{\textbf{ICEWS18}} & \multicolumn{3}{c}{\textbf{GDELT}} \\
			\cmidrule(l){4-6} \cmidrule(l){7-9}  \cmidrule(l){10-12}
			&       &  & \textbf{Hits@1} & \textbf{Hits@3} & \textbf{Hits@10} & \textbf{Hits@1} & \textbf{Hits@3} & \textbf{Hits@10} & \textbf{Hits@1} & \textbf{Hits@3} & \textbf{Hits@10} \\
			\midrule
			\textbf{Embed-based} & xERTE& $\boldsymbol{\circ}$ & 0.330  & 0.454 & 0.570  & 0.209 & 0.335 & 0.462 & 0.085 & 0.159 & 0.265 \\
			& RE-GCN & $\boldsymbol{\circ}$ & 0.313 & 0.473 & 0.626 & 0.223 & 0.367 & 0.525 & 0.084 & 0.171 & 0.299 \\
			\midrule
			\textbf{Rule-based} & TLogic & $\boldsymbol{\circ}$ & 0.332 & 0.476 & 0.602 & 0.204 & 0.336 & 0.48  & 0.113 & 0.212 & \textbf{0.351} \\
			\midrule
			{\textbf{LLM-based}} & InternLM2-7B-ONSEP & \XSolidBrush & 0.330  & 0.464 & 0.570  & 0.200   & 0.324 & 0.443 & - & - & - \\
			\midrule
			\midrule
			\multirow{16}{2cm}{\textbf{LLMs w/wo SPARK}} & InternLM2-ICL & \XSolidBrush & 0.301 & 0.432 & 0.56  & 0.172 & 0.289 & 0.434 & 0.084 & 0.152 & 0.254 \\
			& InternLM2-ICL+SPARK(R) & $\boldsymbol{\circ}$ & 0.351 & 0.493 & 0.601 & 0.197 & 0.328 & 0.435 & 0.108 & 0.191 & 0.281 \\
			& InternLM2-ICL+SPARK(G) & $\boldsymbol{\circ}$ & 0.359 & 0.507 & 0.624 & 0.202 & 0.339 & 0.482 & 0.109 & 0.194 & 0.302 \\
			&\textbf{$\triangle$ \textit{Improve (R)}} & &16.61\% & 14.12\% & 7.32\% & 14.53\% & 13.49\% & 0.23\% & 28.57\% & 25.66\% & 10.63\% \\
			&\textbf{$\triangle$ \textit{Improve (G)}} & &19.27\% & 17.36\% & 11.43\% & 17.44\% & 17.30\% & 11.06\% & 29.76\% & 27.63\% & 18.90\% \\
			
			\cmidrule{2-12}          & Llama2-7B-ICL & \XSolidBrush & 0.252 & 0.427 & 0.504 & 0.127 & 0.272 & 0.323 & 0.06  & 0.164 & 0.246 \\
			& Llama2-7B-ICL + SPARK(R) & $\boldsymbol{\circ}$ & 0.366 & 0.515 & 0.616 & 0.219 & 0.352 & 0.476 & 0.117 & 0.21  & 0.307 \\
			& Llama2-7B-ICL+ SPARK(G) & $\boldsymbol{\circ}$ & 0.362 & \underline{0.517} & 0.635 & 0.21  & 0.356 & \underline{0.499} & 0.118 & 0.211 & 0.326 \\
			&\textbf{$\triangle$ \textit{Improve (R)}}&  & 45.24\% & 20.61\% & 22.22\% & 72.44\% & 29.41\% & 47.37\% & 95.00\% & 28.05\% & 24.80\% \\
			&\textbf{$\triangle$ \textit{Improve (G)}}&  & 43.65\% & 21.08\% & 25.99\% & 65.35\% & 30.88\% & 54.49\% & 96.67\% & 28.66\% & 32.52\% \\
			
			\cmidrule{2-12}          & GPT-NeoX-20B-TLR & \XSolidBrush & 0.35  & 0.474 & 0.575 & 0.211 & 0.339 & 0.456 & 0.102 & 0.167 & 0.273 \\
			& GPT-NeoX-20B-TLR + SPARK(R) & $\boldsymbol{\circ}$ & 0.361 & 0.495 & 0.605 & 0.209 & 0.338 & 0.451 & 0.123 & 0.218 & 0.305 \\
			& GPT-NeoX-20B-TLR + SPARK(G) & $\boldsymbol{\circ}$ & \textbf{0.372} & 0.517 & \textbf{0.642} & 0.215 & 0.35  & 0.495 & 0.124 & 0.216 & 0.331 \\
			&\textbf{$\triangle$ \textit{Improve (R)}} &   &  3.14\% & 4.43\% & 5.22\% & -0.95\% & -0.29\% & -1.10\% & 20.59\% & 30.54\% & 11.72\% \\
			&\textbf{$\triangle$ \textit{Improve (G)}} &   &  6.29\% & 9.07\% & 11.65\% & 1.90\% & 3.24\% & 8.55\% & 21.57\% & 29.34\% & 21.25\% \\
			
			\cmidrule{2-12}          & Llama2-7B-TLR + IT & \Checkmark & 0.369 & 0.475 & 0.532 & \textbf{0.242} & \underline{0.372} & 0.421 & \textbf{0.139} & \underline{0.225} & 0.304 \\
			& Llama2-7B-TLR + SPARK(R) & $\boldsymbol{\circ}$ & 0.358 & 0.504 & 0.609 & 0.222 & 0.356 & 0.478 & 0.13  & 0.222 & 0.319 \\
			& Llama2-7B-TLR + SPARK(G) & $\boldsymbol{\circ}$ & \underline{0.371} & \textbf{0.520} & \underline{ 0.638} & \underline{0.233} & \textbf{0.373} & \textbf{0.509} & \underline{0.131} & \textbf{0.229} & \underline{0.350} \\
			\bottomrule
		\end{tabular}%
	}
	\label{tab:main}
\end{table*}

\subsection{Experimental Analysis}
\subsubsection{Main Results (RQ1)}
From Table \ref{tab:main}, we observe:
\textbf{(1) Effectiveness of SPARK:} 
SPARK consistently enhances LLM performance, especially making smaller 7B models competitive, demonstrating its cost-effective reasoning enhancement.
\textbf{(2) SPARK vs. Traditional Methods:} 
SPARK outperforms traditional methods like TLogic and xERTE by intelligently combining their strengths with LLMs. It adapts to when traditional models excel and when LLMs perform better.
\textbf{(3) SPARK vs. IT:}
SPARK achieves comparable performance to IT-tuned models with fewer resources, eliminating costly fine-tuning and manual instructions.
\textbf{(4) SPARK(R) vs. SPARK(G):} 
SPARK(G) excels SPARK(R) (especially in Hits@10) due to xERTE’s large-scale neighborhood exploration and larger adaptable parameter space, while SPARK(R) with TLogic provides explainable symbolic reasoning but captures fewer broad patterns.

\subsubsection{Generalization Ability (RQ2)}
\begin{figure}\centering
	\includegraphics[width=0.65\linewidth]{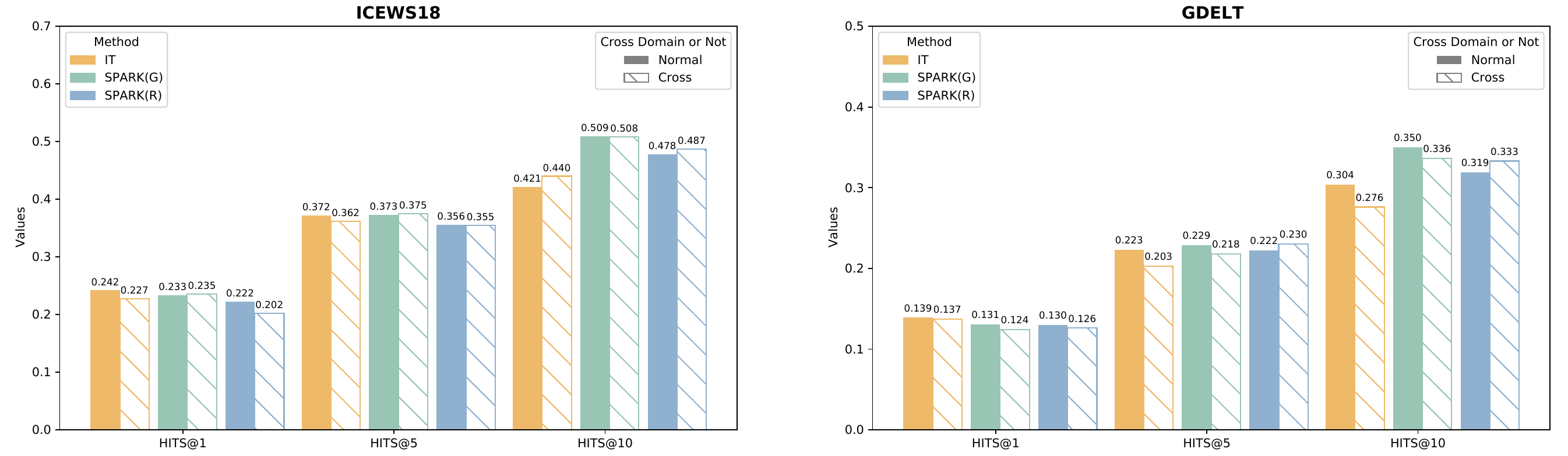}
	\caption{Results of cross-domain generalizability experiments. The Cross ones (shaded) are trained on ICEWS14, while the Normal ones (solid) are trained on specific datasets.}
	\label{fig_inductive_across}
\end{figure}

One of the most significant advantages of LLMs is their cross-domain generalization ability. To assess this, we compared SPARK with IT. To be specific, we trained Llama2-7B-TLR on ICEWS14 and evaluated the models on ICEWS18 and GDELT. As shown in Figure \ref{fig_inductive_across}, both SPARK(R)-Cross and SPARK(G)-Cross perform comparably to models trained on target datasets. SPARK also exhibits smaller performance drops than IT in Hits@3 and Hits@10, demonstrating stronger robustness.

\subsubsection{Efficiency (RQ3)}
SPARK is a cost-effective method for refining LLMs for TKG forecasting. To demonstrate this, we compare SPARK(G) and SPARK(R) with GenTKG \cite{gentkg}, which uses IT (LoRA rank of 8) and an instruction-based iterative K times generation. As shown in Table \ref{tab:efficency}, SPARK reduces training and inference time per iteration, ensuring high efficiency. This is achieved through three key factors: 1) Its instruction-free design halves prompt length, 2) As an end-to-end inference-time adapter, LLM results can be pre-stored when training, and 3) BSL generation is more efficient than iterative top-K generation.
\begin{table}
    \centering
    \caption{Efficiency analysis on ICEWS14. We report the time per batch (s/it) using a Nvidia A100 GPU. Note that the inference time is the sum of LLM and the adapter.}
    {\fontsize{7pt}{8pt}\selectfont
    \resizebox{0.99\textwidth}{!}{
    \begin{tabularx}{\textwidth}{l|X|X|X}
    \toprule
    \textbf{Method}        & \textbf{Prompt Length} & \textbf{Training Time} & \textbf{Inference Time} \\
    \midrule
    \textbf{GenTKG (IT, Iterative K)}      & 2435.3                    & 75.86                      & 72.41                          \\ 
    \textbf{SPARK (G)}      & 1389.4                    & 2.11                       & 19.62              \\ 
    \textbf{SPARK (R)}   & 1389.4                    & 13.37                      & 29.24                          \\  \bottomrule
    \end{tabularx}
}}
    \label{tab:efficency}
\end{table}

\subsubsection{Ablation Study (RQ4)}
We conduct ablation studies to assess each component’s contribution, with results shown in Table \ref{tab:ablation}. For the BSL Generation ablation, we replace it with iterative top-K generation \cite{gentkg}. We found that BSL generation significantly boosts performance across all datasets, making it potentially useful for applications requiring top-K generation. TKGAdapter primarily improves HITS@10, emphasizing its role in refining predictions through broader structural reasoning.
\begin{table}
\centering
\caption{Results of ablation study for SPARK.}
{\fontsize{7pt}{8pt}\selectfont
\resizebox{0.99\textwidth}{!}{
\begin{tabularx}{\textwidth}{l|XXXXXX}
\toprule
\multirow{2}{*}{\textbf{Model}} & \multicolumn{2}{c}{\textbf{ICEWS14}} & \multicolumn{2}{c}{\textbf{ICEWS18}} & \multicolumn{2}{c}{\textbf{GDELT}} \\ 
\cmidrule(l){2-3} \cmidrule(l){4-5}  \cmidrule(l){6-7}
 & \textbf{Hit@1} & \textbf{Hit@10} & \textbf{Hit@1} & \textbf{Hit@10} & \textbf{Hit@1} & \textbf{Hit@10} \\ 
\midrule
\textbf{SPARK(G)} & 0.371 & 0.638 & 0.233 & 0.509 & 0.131 & 0.350 \\ 
\textbf{- BSL Generation} & 0.249 & 0.561 & 0.185 & 0.439 & 0.104 & 0.298 \\ 
\textbf{- TKGAdapter} & 0.368 & 0.601 & 0.218 & 0.458 & 0.128 & 0.316 \\ 
\bottomrule
\end{tabularx}
}}
\label{tab:ablation}
\end{table}

\section{Conclusion And Discussion}
We introduce SPARK, a sequence-level generation and proxy adapting framework for refining LLMs for TKG forecasting at inference time. By leveraging beam sequence-level generation and TKG-specific adapters, SPARK enhances predictive accuracy and efficiency, making it a practical solution for real-world applications. Looking ahead, SPARK holds great potential for expanding into multi-step temporal reasoning with reinforcement learning.

\begin{credits}
\subsubsection{\ackname}
This work was supported by the Natural Science Foundation of Heilongjiang Province of China (Grant No. LH2023F018), and the Fundamental Research Funds for the Central Universities (Grant No. LH2023F018).
\end{credits}

\bibliographystyle{splncs04}
\bibliography{samplepaper}
\end{document}